\documentclass[runningheads]{llncs}
\usepackage{graphicx}

\usepackage[hang,flushmargin]{footmisc}
\usepackage[misc,geometry]{ifsym}
\usepackage{cite} 
\usepackage[T1]{fontenc}
\usepackage{amsmath}
\usepackage{mathrsfs}
\usepackage{graphicx}  
\usepackage{epstopdf}
\usepackage{bbm}
\usepackage{float}
\usepackage{multirow}
\usepackage{booktabs}
\usepackage{mathrsfs}
\usepackage{amssymb}
\usepackage[colorlinks,linkcolor=red]{hyperref}
\usepackage{siunitx}
\makeatletter

\newcommand{\Rmnum}[1]{\expandafter\@slowromancap\romannumeral #1@}
\makeatother
\usepackage{booktabs}
\usepackage{amsfonts,amssymb}
\usepackage{amsmath,bm}
\usepackage{array}
\newcommand{\PreserveBackslash}[1]{\let\temp=\\#1\let\\=\temp}
\newcolumntype{C}[1]{>{\PreserveBackslash\centering}p{#1}}
\newcolumntype{R}[1]{>{\PreserveBackslash\raggedleft}p{#1}}
\newcolumntype{L}[1]{>{\PreserveBackslash\raggedright}p{#1}}
\usepackage{algorithmicx,algorithm}
\makeatletter

\newcommand{\para}[1]{\vspace{.05in}\noindent\textbf{#1}}

\usepackage{xcolor}

\begin{document}

\title{EndoGS: Deformable Endoscopic Tissues Reconstruction with Gaussian Splatting}

\titlerunning{EndoGS}

\author{
Lingting Zhu\inst{1}\and
Zhao Wang\inst{2} \and
Jiahao Cui\inst{3} \and 
Zhenchao Jin\inst{1} \and \\
Guying Lin\inst{1} \and 
Lequan Yu\inst{1}\textsuperscript{(\Letter)}
}


\authorrunning{L. Zhu et al.}

%
\institute{
    The University of Hong Kong, Hong Kong SAR, China \\ \email{ltzhu99@connect.hku.hk lqyu@hku.hk} \and
    The Chinese University of Hong Kong, Hong Kong SAR, China \\ \and
    Sun Yat-sen University, Guangzhou, China
}

\maketitle

\begin{abstract}

Surgical 3D reconstruction is a critical area of research in robotic surgery, with recent works adopting variants of dynamic radiance fields to achieve success in 3D reconstruction of deformable tissues from single-viewpoint videos. However, these methods often suffer from time-consuming optimization or inferior quality, limiting their adoption in downstream tasks.
Inspired by 3D Gaussian Splatting, a recent trending 3D representation, we present \textbf{EndoGS}, applying Gaussian Splatting for deformable endoscopic tissue reconstruction.
Specifically, our approach incorporates deformation fields to handle dynamic scenes, depth-guided supervision with spatial-temporal weight masks to optimize 3D targets with tool occlusion from a single viewpoint, and surface-aligned regularization terms to capture the much better geometry.
As a result, EndoGS reconstructs and renders high-quality deformable endoscopic tissues from a single-viewpoint video, estimated depth maps, and labeled tool masks.
Experiments on DaVinci robotic surgery videos demonstrate that EndoGS achieves superior rendering quality.
Code is available at {\href{https://github.com/HKU-MedAI/EndoGS}{https://github.com/HKU-MedAI/EndoGS}}.


\end{abstract}
\section{Introduction}
\label{sec:intro}

Reconstruction of high-quality deformable tissues from endoscopic videos is a significant but challenging task, facilitating downstream tasks like surgical AR, education, and robot learning~\cite{chen2018slam, scott2008changing, shin2019autonomous}. Earlier attempts~\cite{lu2021super, recasens2021endo, song2017dynamic, zhou2019real, zhou2021emdq} adopt depth estimation to achieve great success in endoscopic reconstruction, but these methods still struggle to produce realistic 3D reconstruction due to two key issues. First, non-rigid deformations with sometimes large movements, which require practical dynamic scene reconstruction, hinders the adaptation of those techniques. Second, tool occlusion happens in single-viewpoint videos, producing difficulties in learning affected parts with limited information.

While~\cite{li2020super, long2021dssr} have proposed frameworks combining tool masking, stereo depth estimation, and sparse warp field~\cite{gao2019surfelwarp, newcombe2015dynamicfusion} for single-viewpoint 3D reconstruction, they are still prone to failure in the presence of dramatic non-topological deformable tissues changes. Neural Radiance Fields (NeRFs)~\cite{mildenhall2021nerf} have shown great potentials in 3D representations, and methods based on variants of dynamic radiance field~\cite{pumarola2021d, fridovich2023k} have become representative works in deformable tissues reconstruction from videos. For example, EndoNeRF~\cite{wang2022neural} follows the modeling of D-NeRF~\cite{pumarola2021d} to represent deformable surgical scenes as the combination of a canonical neural radiance field and a time-dependent neural displacement field, and LerPlane~\cite{yang2023neural} factorizes six 2D planes for static field and dynamic field to accelerate optimization similar to~\cite{fridovich2023k, cao2023hexplane}.

Recently 3D Gaussian Splatting (3D-GS)~\cite{kerbl20233d, wu2024recent, chen2024survey} has been witnessed as a powerful representation that renders higher-quality results at a real-time level. Follow-up works~\cite{luiten2023dynamic, yang2023deformable, wu20234d, yang2023gs4d} extend 3D-GS to represent dynamic scenes and achieve state-of-the-art performances on rendering fidelity and speed. To model dynamic representation of dynamic scenes,~\cite{luiten2023dynamic, yang2023gs4d} formulate 4D Gaussians and assign extra parameters as attributes in the time dimension, and~\cite{wu20234d, yang2023deformable} apply MLPs to predict the deformation of the Gaussians, sharing the same rationale with dynamic NeRFs~\cite{park2021nerfies, pumarola2021d}. In this paper, we present EndoGS, a method based on surface-aligned Gaussian Splatting for deformable endoscopic tissues reconstruction with better rendering quality and better rendering speed. 

To summarize, our main contributions are three-fold: \textbf{1)} We present the first Gaussian Splatting based method for deformable endoscopic tissues reconstruction. This is one of the first attempts~\cite{li2023sparse} introducing Gaussian Splatting in the medical domain. \textbf{2)} We represent dynamic surgical scenes with the combination of static Gaussians and the deformable parameters in the time dimension, adopt depth-guided supervision with spatial-temporal weight masks for monocular optimization with tool occlusion, and combine surface-aligned regularization terms ~\cite{guedon2023sugar} to capture the much better geometry. \textbf{3)} We use the same input tool masks involved in the training and inference for comparison methods and make a clear and fair comparison, and experiments demonstrate our superior performances.

\section{Method}

\begin{figure}[t]
\includegraphics[width=1.0\textwidth, height=0.45\textwidth]{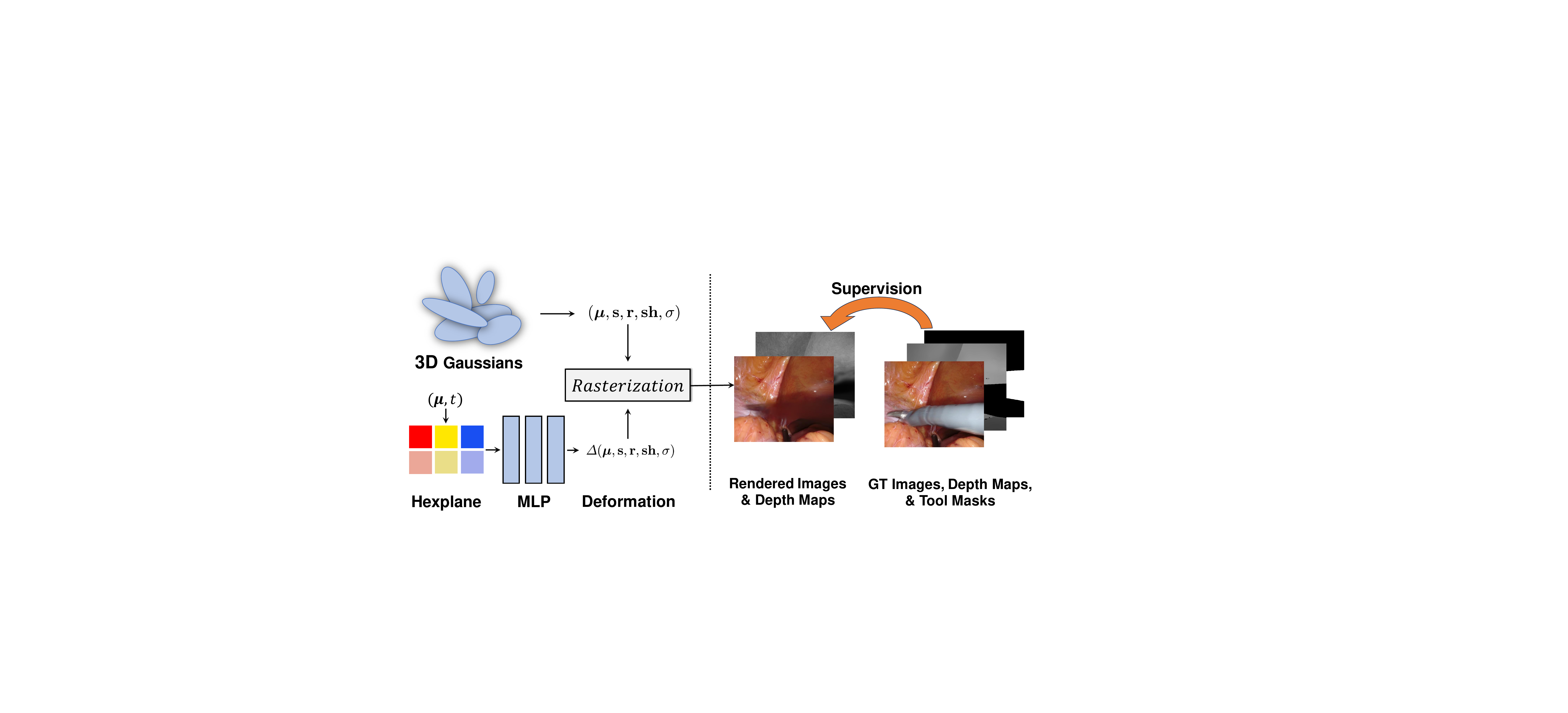}
\caption{\textbf{The overview of our EndoGS pipeline.} 
Given 3D Gaussians, we use the mean and the time as input to compute features by querying multi-resolution voxel planes. A single MLP is used to obtain the deformation of the Gaussians. With differentiable rasterization, the rendered images and depth maps are obtained and we use ground truth images, depth maps and the tool masks to provide the supervision.} 
\label{overview}
\end{figure}

\subsection{Overview}
In this paper, we introduce our method, referred to as EndoGS, which utilizes a deformable variant of 3D-GS to reconstruct 3D surgical scenes from a single-viewpoint video, estimated depth maps, and labeled tool masks. Specifically, given a stereo video with left and right frames $\{(\bm{I}_i^l, \bm{I}_i^r)\}_{i=1}^T$, where $T$ is the total number of frames, our goal is to reconstruct 3D representations of the deformable tissues that render in high quality. We follow the approach of~\cite{wang2022neural, yang2023neural, yang2023efficient} by combining the extracted binary tool masks $\{\bm{M}_i\}_{i=1}^{T}$ and the depth maps $\{\bm{D}_i\}_{i=1}^{T}$ estimated from binocular captures for the left views. The pipeline of EndoGS is illustrated in Fig.~\ref{overview}. In this section, we first introduce the preliminary of 3D Gaussian Splatting (Section~\ref{sec pre}). We then present the modeling of deformable tissues with a dynamic version of 3D-GS, which adopts a lightweight MLP to represent the dynamic field (Section~\ref{sec 4dgs}). The training optimization of the Gaussian Splatting with the tool masks and the depth maps is introduced (Section~\ref{sec tool depth}). Finally, we discuss how to align Gaussians with the Surface (Section~\ref{sec sugar}).
 
\subsection{Preliminaries of 3D Gaussian Splatting}
\label{sec pre}
3D Gaussian Splatting (3D-GS) \cite{kerbl20233d} offers the state-of-the-art solution for real-time novel view synthesis in multi-view static scenes. 
The 3D Gaussians with a 3D covariance matrix $\mathbf{\Sigma}$ and mean $\bm \mu$ served as the rendering primitives in the form of point clouds:
\begin{equation}
    G(\bm{x})=e^{-\frac{1}{2}(\bm{x}-\bm \mu)^T\mathbf{\Sigma}^{-1}(\bm{x-\bm \mu})}.
\end{equation}
The 3D Gaussians can be projected onto 2D space and rendered for pixels: $\mathbf{\Sigma}^{\prime} = \mathbf{J}\mathbf{W}\mathbf{\Sigma} \mathbf{W}^T\mathbf{J}^T$, where $\mathbf{\Sigma}^{\prime}$ is the covariance matrix in the 2D plane, $\mathbf{W}$ denotes view transformation, and the Jacobian $\mathbf{J}$ is the affine approximation of the projective transformation.
To enforce the positive semi-definiteness, $\mathbf{\Sigma}$ is parameterized as $\mathbf{R}$: $\mathbf{\Sigma} = \mathbf{R}\mathbf{S}\mathbf{S}^T\mathbf{R}^T$ with a scale $\mathbf{S}$ and rotation.

To render the color of the pixels $\bm{p}$, point-based volume rendering is adopted:
\begin{align}
C(\bm{p}) = \sum_{i\in N}c_i \alpha_i \prod_{j=1}^{i-1} (1-\alpha_j),\\
\textit{where } \alpha_i = \sigma_i e^{-\frac{1}{2}(\bm{p-\bm \mu_i})^T{\mathbf{\Sigma}^{\prime}}^{-1}(\bm{p-\bm \mu_i})}.
\end{align} 
The $c_i$ means the color of the Gaussians along the ray, $\bm \mu_i$ denotes the coordinates, and $\sigma_i$ denotes the opacity.
And 3D-GS applies spherical harmonics to model the view-dependent color. 

In total, the explicit 3D Gaussians are characterized by: position $\bm{\mu} \in \mathbb{R}^3$, scaling factor $\bm{s} \in \mathbb{R}^3$, rotation factor $\bm{r} \in \mathbb{R}^4$, spherical harmonic (SH) coefficients $\bm{sh} \in \mathbb{R}^k$ ($k$ means number of SH functions), and opacity $\sigma \in \mathbb{R}$. Finally each Gaussian can be represented as ($\bm \mu, \bm{s}, \bm{r}, \bm{sh}, \sigma$).

\subsection{Gaussian Splatting Representations for Deformable Tissues}
\label{sec 4dgs}

We represent a surgical scene as a 4D volume, where the deformation of the tissues involves time dimension. To this end, we introduce the Gaussian deformation to represent the time-varying motions and shapes, following the basic designs of~\cite{wu20234d}. The final goal is to learn the original representation of the 3D Gaussians $\{(\bm \mu, \bm s, \bm r, \bm{sh}, \sigma)\}$ as well as the Gaussian deformations $\{\Delta (\bm \mu, \bm s, \bm r, \bm {sh}, \sigma)\} = \{(\Delta \bm \mu, \Delta \bm s, \Delta \mathbf r, \Delta \bm {sh}, \Delta \sigma) \}$.

In Fig.~\ref{overview}, for each 3D Gaussian, we use the mean ${\bm \mu} = (x,y,z)$ and the time $t$ to compute the deformation. We use six orthogonal feature planes to encode the spatial and temporal information~\cite{fridovich2023k, cao2023hexplane, yang2023neural, yang2023efficient, wu20234d}. To be specific, the multi-resolution HexPlane~\cite{cao2023hexplane, fridovich2023k} consists of three space planes $XY, XZ, YZ$ and three space-time planes $XT, YT, ZT$. The planes encode features $F \in \mathbb{R}^{h\times N_1 \times N_2}$, where $h$ denotes the hidden dimension and $N_1, N_2$ stand for the plane resolution, and we utilize bilinear interpolation $\mathcal{B}$ to interpolate the four nearby queried voxel features.
As a result, the voxel feature can be represented in the format of matrix element-wise multiplication with operation $\odot$:
\begin{align}
f_{voxel}({\bm \mu}, t) = \mathcal{B}\left(F_{X Y}, x, y\right) \odot \mathcal{B}\left(F_{Y Z}, y, z\right) \ldots \mathcal{B}\left(F_{Y T}, y, t \right) \odot \mathcal{B}\left(F_{Z T}, z, t \right).
\end{align}
Then we employ a single MLP to update the Gaussian attributes and it merges all the information and decodes the deformation of the position, scaling factor, rotation factor, spherical harmonic coefficients, and opacity:
\begin{align}
\Delta {(\bm \mu, \bm s, \bm r, \bm{sh}, \sigma)} = {\rm MLP}(f_{voxel}({\bm \mu}, t)).
\end{align}

\subsection{Training Combined with Tool Masks and Depth Maps}
\label{sec tool depth}
Reconstructing from videos with tool occlusion poses a challenge and we follow former works~\cite{wang2022neural, yang2023neural, yang2023efficient} to use labeled tool occlusion masks to indicates the unseen pixels. Furthermore, we leverage spatiotemporal importance sampling strategy to indicate the crucial areas related to the occlusion issue. To be specific, the binary masks are denoted as $\{\bm{M}_i\}^T_{i=1}$, where 0 stands for tissue pixels and 1 stands for tool pixels, and the importance maps $\{\bm{\mathcal{V}}_i\}^T_{i=1}$ involve temporal statistics and are denoted as
$\bm{\mathcal{V}}_i = (\bm{1}-\bm{M}_i)\odot \left(\bm{1}+\alpha \sum_{j=1}^T \bm{M}_j /\left\|\sum_{j=1}^T \bm{M}_j\right\|_F\right).$
We only optimize in the seen part by introducing the item $\bm{1}-\bm{M}_i$. Meanwhile, the statistics of the mask frequencies normalized by the Frobenius norm along the temporal dimension provide the information of the uncertainty and allocate higher importance for tissue areas with higher occlusion frequencies. The parameter $\alpha$ is used to control the scaling strength. We use the $L_1$ loss with spatial masks, and thus the spatial supervision on $i$-th is
\begin{align}
\mathcal{L}_{L1}(i) = |\bm{I}_i \odot \bm{\mathcal{V}}_i - \hat{\bm I}_i \odot \bm{\mathcal{V}}_i|,
\end{align}
where $\hat{\bm I}_i$ is the rendering image on $i$-th frame.

Monocular reconstruction provides limited information for 3D reconstruction and makes overfitting happen with single-viewpoint images~\cite{chung2023depth}. To mitigate illness from single-viewpoint inputs, we introduce depth-guided loss with the estimated depth maps. The coarse stereo depth maps are obtained via STTR-light~\cite{li2021revisiting}. We adopt Huber loss $\mathcal{L}_D(i)$ for depth regularization following~\cite{yang2023neural}.

We adopt total variation (TV) losses in the spatial dimension and the temporal dimension to serve as the additional regularization. To prevent color drifts in unseen area, we use a spatial total variation item for areas with tool masks and it is denoted as $\mathcal{L}_{tv-spatial}(i) = {\rm TV} (\hat{\bm{I}}_i \odot \bm{\mathcal{M}}_i)$. And we use the temporal total variation item $\mathcal{L}_{tv-temporal}$ in~\cite{wu20234d} to regularize the Hexplane optimization.

\subsection{Surface-Aligned Gaussians}
\label{sec sugar}
When reconstructing the region with limited 3D cues, especially for surrounding tool occlusion masks, noticeable artifacts on the surface are evident.~\cite{guedon2023sugar}  proposes a technique aimed at facilitating precise and rapid mesh extraction from 3D-GS.

To ensure tight integration of the Gaussians with the surface, surface alignment normalization is applied by controlling the density function of the Gaussians, which can be defined as
$ d(\bm p) = \sum_{g} \sigma_g \exp\left(-\frac{1}{2}(\bm p - \bm \mu_g)^T \mathbf{\Sigma}^{-1}_g (\bm p - \bm \mu_g)\right) \>$, where $\bm p$ denotes the position and $g$ denotes the Gaussian.

If the Gaussians are aligned with the surface, we could have three assumptions. 1) The closest Gaussian $g^* = \arg\min_g \left\{ (\bm p - \bm \mu_g)^T \mathbf{\Sigma}^{-1}_g (\bm p - \bm \mu_g) \right\} \>$ to the point $\bm p$ is likely to contribute much more than others to the density value $d(\bm p)$. 2) To ensure the 3D Gaussians to be flat, every Gaussian $g$ would have one of its three scaling factors close to zero. 3) Gaussians are opaque and we can encourage $\sigma_i$ to be 1 with cross-entropy loss.
Under these assumptions, the density can be approximated as 
$ \bar{d}(\bm p) = \exp\left(-\frac{1}{2s_{g^*}^2} \langle \bm p - \bm \mu_{g^*}, \bm n_{g^*} \rangle^2 \right) \>$, where $s_{g^*}$ is the smallest value of $\bm s_{g^*}$ and $\bm n_{g^*}$ is the direction of the corresponding axis. Based on the approximated expression of $\bar d(\bm p)$, we could find the zero-crossings of
the Signed Distance Function and approximate the ideal distance function associated with the density function $d$ as 
$f(\bm p) = \pm s_{g*} \sqrt{-2 \log \left( d(\bm p) \right)}.$
We do not use the density loss in the implementation of~\cite{guedon2023sugar} and refer to~\cite{guedon2023sugar} for details of the approximation.

We use the regularization term on the ideal SDF $f(\bm p)$ and an estimated $\hat f(\bm p)$ that is the difference between the depth
of $\bm p$ and the depth in the corresponding depth map at the
projection of $\bm p$:
\begin{align}
\mathcal{L}_{SDF} = \frac{1}{|\mathcal{P}|} \sum_{\bm p\in\mathcal{P}} |\hat{f}(\bm p) - f(\bm p)|.
\end{align}
The estimated $\hat f(\bm p)$ is defined as the difference between the depth of $\bm p$ and the depth in the corresponding depth map at the projection of $\bm p$. Meanwhile, we add a regularization term for normal direction:
\begin{equation}
    \mathcal{L}_{norm} = \frac{1}{|\mathcal{P}|} \sum_{p\in\mathcal{P}}
    \left\| \frac{\nabla f(\bm p)}{\|\nabla f(\bm p)\|_2} -  \bm n_{g^*}\right\|_2^2 \>.
\end{equation}
Besides, to encourage opaque Gaussians, we adopt cross-entropy loss for opacity regularization:
\begin{align}
\mathcal{L}_{opacity}(i) = -\sum_{j} (\sigma + \Delta \sigma)_{j}\log (\sigma + \Delta \sigma)_{j},
\end{align}
where $(\sigma + \Delta \sigma)_{j}$ denotes the $j$-th Gaussian's opacity in the $i$-th frame.

To sum up, our final optimization target at the $i$-th frame is:
\begin{equation}
    \mathcal{L}(i) = \mathcal{L}_{L1}(i) + \lambda_D \mathcal{L}_D(i) + \lambda_{TV1} \mathcal{L}_{tv-spatial}(i) + \lambda_{TV2} \mathcal{L}_{tv-temporal} 
    + \lambda_{S} \mathcal{L}_{S}(i),
    \label{eq:R_norm}
\end{equation} where $\mathcal{L}_{S}(i) = \mathcal{L}_{SDF}+\mathcal{L}_{norm}+0.5\mathcal{L}_{opacity}(i)$ denotes the surface-aligned regularization. Hyperparameters $\lambda_D$, $\lambda_{TV1}$, $\lambda_{TV2}$, $\lambda_{S}$ control the strength. 
\section{Experiments}

\subsection{Datasets and Evaluation Metrics}
We evaluate our proposed method on the dataset from~\cite{wang2022neural}: typical robotic surgery videos from 6 cases of DaVinci robotic data. The datasets are designed to capture challenging surgical scenes with non-rigid deformation and tool occlusion. We use standard image quality metrics, including PSNR, SSIM, and LPIPS. Since in evaluation the groundtruth pixels for unseen areas are missing, the tool masks are used to exclude unseen parts for computation and unlike~\cite{wang2022neural, yang2023neural, yang2023efficient}, we do not count those pixels in PSNR. We also report the frame per second (FPS) to compare the rendering speed of the methods.
Besides, while former works~\cite{wang2022neural, yang2023neural, yang2023efficient} adopt the different tool mask configurations in training and evaluation or compare methods under different configurations, we argue to train and evaluate in the same tool mask configurations to prevent meaningless pixels comparison, and compare methods the same tool occlusion masks.

\subsection{Implementation Details}
In our approach, we adopt the two-stage training methodology in~\cite{wu20234d} to model the static and deformation fields. In the first stage, we train 3D Gaussian models for the static field, while in the second stage, we focus on training the deformation field. We conduct 3,000 and 60,000 iterations for the first and second stages, respectively. The initial point clouds are estimated using COLMAP~\cite{schonberger2016structure}. The importance maps scaling strength $\alpha$ is set to 30 and $\delta$ = 0.2 for depth loss. Hyperparameters $\lambda_D$, $\lambda_{TV1}$, $\lambda_{TV2}$, $\lambda_{S}$ are set to 0.5, 0.01, 1.0, 0.2. We train our models on an NVIDIA RTX 3090 GPU.

\subsection{Results}

\begin{table*}[t]
  \centering
  \caption{\textbf{Quantitative comparison on rendering quality and speed (standard deviation in parentheses).}}
  \begin{tabular}{l|c|c|c|ccccc}
    \toprule
    Methods & PSNR $\uparrow$ & SSIM $\uparrow$ & LPIPS $\downarrow$ & FPS$\uparrow$ \\
    \hline
   EndoNeRF~\cite{wang2022neural}
    & 35.112 (1.470) & 0.936 (0.021) & 0.066 (0.030) &  $< 0.2$ \\
   ForPlane~\cite{yang2023efficient}  & 36.427 (1.214)  & 0.947 (0.007)   & 0.058 (0.003)    &  $\sim 1.7$  \\
   \hline 
    EndoGS w/o SA & 37.603 (0.076)  & 0.964 (0.005) & 0.036 (0.009) &  $\sim 70$ 
    \\
    \textbf{EndoGS (Ours)} & \textbf{37.935(0.088)}  & \textbf{0.966(0.003)} & \textbf{0.034(0.001)} &  $\sim 70$ 
    \\
    \bottomrule[1pt]
  \end{tabular}
  \label{tbl:res}
\end{table*}

\begin{figure}[t]
\vspace{-5pt}
\includegraphics[width=\textwidth]{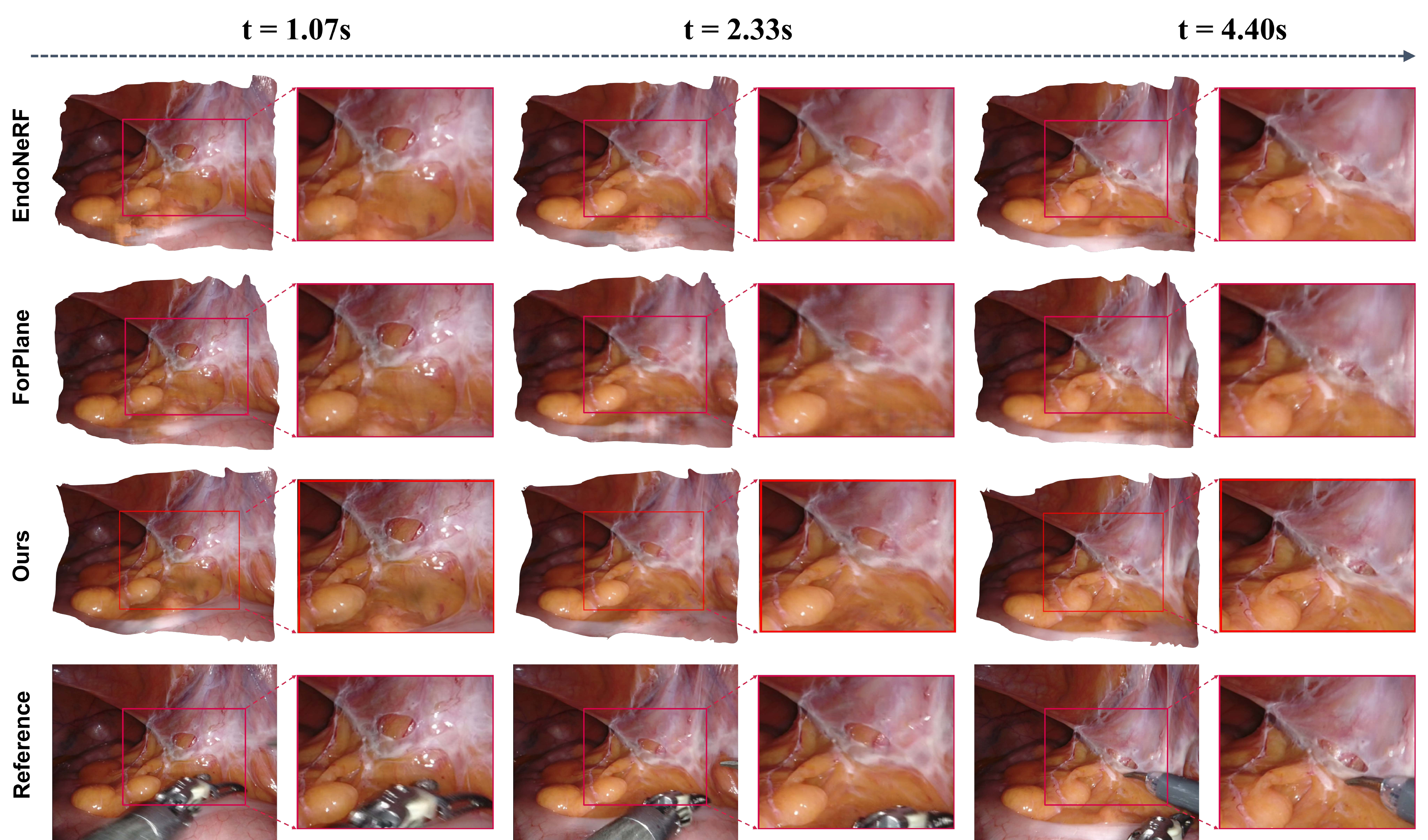}
\caption{\textbf{Qualitative results on scene “traction” at different timesteps.}} 
\label{results}
\end{figure}

We compare EndoGS against two methods, \textit{i.e.,} EndoNeRF~\cite{wang2022neural} and ForPlane~\cite{yang2023efficient} (an updated version of LerPlane~\cite{yang2023neural}), due to their competitive quality. ForPlane is trained for 32k iterations. We use the same masks in training and evaluation for three methods and evaluate the rendering results on the same cropped zone where the lowest part of the videos that contain display patterns are removed. 

\begin{figure}[h]
\includegraphics[width=1.0\textwidth]{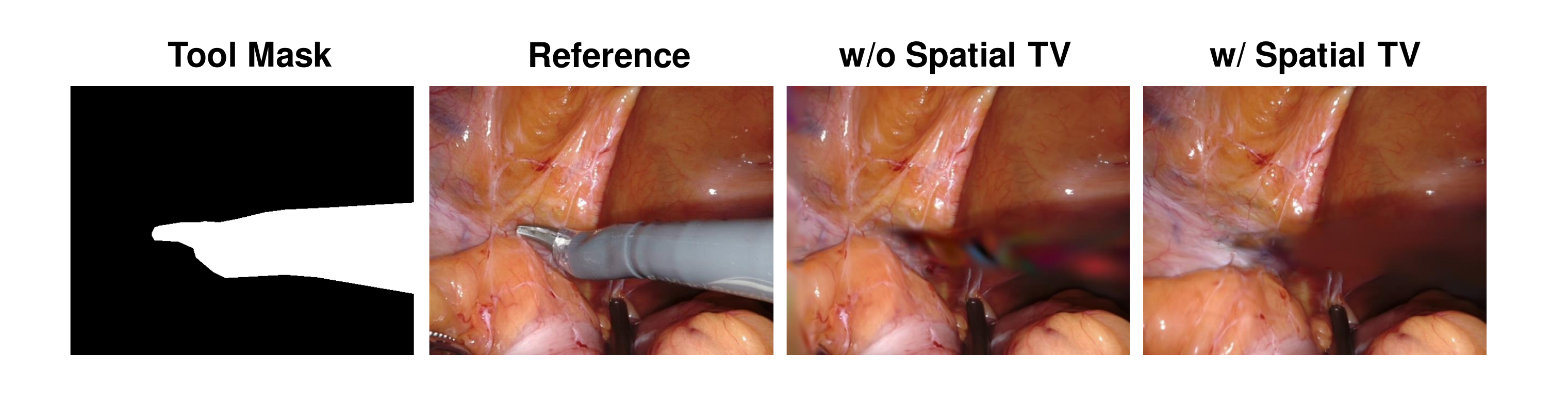}
\caption{\textbf{Ablation on spatial TV loss.} We show rendering frames w/ and w/o spatial TV loss on scene "cutting tissues twice".}
\label{tv}
\end{figure}

\begin{figure}[h]
\includegraphics[width=1.0\textwidth]{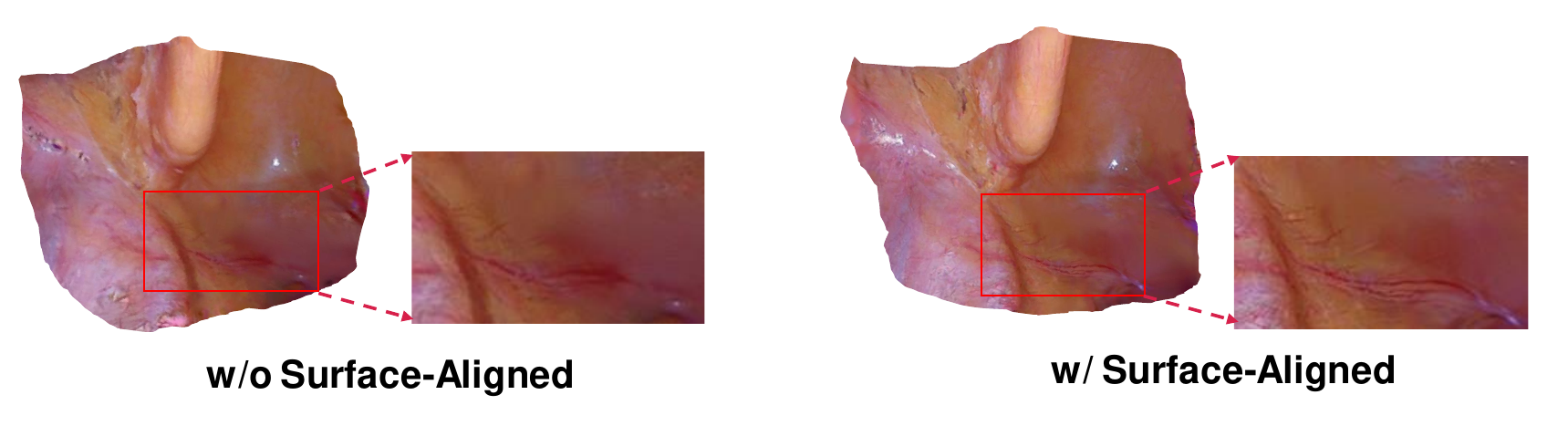}
\caption{\textbf{Ablation on Surface-Aligned regularization.} We show reconstruction results w/o and w/ Surface-Aligned regularization on scene "pulling soft tissues".}
\label{surface}
\end{figure}

Fig.~\ref{results} presents a qualitative comparison on scene "traction" between EndoGS and competitive baselines. On the rendering quality, EndoGS clearly outperforms other methods. Tissues deformation occurs at different timesteps, and our method better reconstructs the deformation along the time. We also include the ablation results over the Surface-Aligned regularization which is denoted as SA in the table. Table~\ref{tbl:res} presents a quantitative comparison on the 6 videos, showcasing the superior performances of our method over the baseline methods in terms of various evaluation metrics related to rendering quality and speed. Considering that EndoNeRF exhibits poor performance on the first rendering frame, we also exclude the first frames from the evaluation process. In terms of rendering quality, EndoGS demonstrates a significant advantage, outperforming the other methods by a considerable margin. Furthermore, EndoGS benefits from the rendering efficiency of Gaussian Splatting, enabling it to achieve real-time rendering speeds. In contrast, the baseline methods struggle to maintain a high FPS rate, highlighting the superiority of EndoGS in this regard.

\para{Ablation Analysis.} 
In Fig.~\ref{tv}, we show the effectiveness of spatial Total Variation loss. While we optimize the Gaussians on seen pixels in frames with the tool masks, the lack of continuity leads to color drift in rendering results (frame w/o spatial TV). This phenomenon can be mitigated during the optimization with spatial TV loss.
Table~\ref{tbl:res} and Fig.~\ref{surface} present the quantitative and qualitative results over the effectiveness of Surface-Aligned regularization. When reconstructing the surrounding of tool occlusion masks, noticeable artifacts on the surface are evident, which cause the rendered image to become blurred. It is shown that with Surface-Aligned regularization more details are preserved.

\section{Conclusion}
We present a method for deformable endoscopic tissue reconstruction based on surface-aligned Gaussian splatting, which achieves high-quality, real-time reconstruction from a single-viewpoint video, estimated depth maps, and labeled tool masks. Experiments on DaVinci robotic surgery videos demonstrate the superior quality of our method. However, there is limitation to 3D reconstruction from single-viewpoint videos, as it is an ill-posed problem that makes it infeasible for surgical downstream applications. For example, we lack 3D cues for tissues that are occluded by tools. Therefore, future work can focus on practical endoscopic reconstruction, and 3D tissue reconstruction with more surgical cameras is needed to facilitate realistic downstream tasks. We would also like to recommend readers referring to other concurrent and follow-up works~\cite{liu2024endogaussian,huang2024endo,bonilla2024gaussian,yang2024efficient,zhao2024hfgs,yang2024deform3dgs,li2024endosparse,liu2024lgs}.

\section*{Acknowledgement} This work was partially supported by the Research Grants Council of Hong Kong ( T45-401/22-N and 27206123) and the National Natural Science Foundation of China (No. 62201483). We thank Med-AIR Lab CUHK for DaVinci robotic prostatectomy data.

\bibliographystyle{splncs04}
\bibliography{reference}

\end{document}